\documentclass[pmlr]{jmlr}%

\usepackage{longtable}%

 \usepackage{booktabs}

\usepackage{bm}
\usepackage{amstext}
\usepackage{tikz,pgfplots}
\usepackage{adjustbox}
 
\usepackage[load-configurations=version-1]{siunitx} %

\makeatletter
\def\set@curr@file#1{\def\@curr@file{#1}} %
\makeatother

\theorembodyfont{\upshape}
\theoremheaderfont{\scshape}
\theorempostheader{:}
\theoremsep{\newline}

\title{Generative models for wearables data}

\author{\Name{Arinbjörn Kolbeinsson}
       \Email{arinbjorn@evidation.com}\\ 
       \addr Evidation Health\\
       San Mateo, CA, USA
       \AND
       \Name{Luca Foschini}
       \Email{luca.foschini@sagebase.org}\\ 
       \addr Sage Bionetworks\\
       Seattle, WA, USA} 

\begin{document}

\maketitle

\begin{abstract}
  Data scarcity is a common obstacle in medical research due to the high costs associated with data collection and the complexity of gaining access to and utilizing data. Synthesizing health data may provide an efficient and cost-effective solution to this shortage, enabling researchers to explore distributions and populations that are not represented in existing observations or difficult to access due to privacy considerations. To that end, we have developed a multi-task self-attention model that produces realistic wearable activity data. We examine the characteristics of the generated data and quantify its similarity to genuine samples with both quantitative and qualitative approaches.
\end{abstract}

\section{Introduction}
\label{sec:intro}
High quality health data is a vital yet scarce resource in modern healthcare. Raw data collection is expensive and time consuming, labelling requires expert knowledge and storage poses privacy concerns. As a result, most health datasets fail to capture the true distribution of the underlying population, particularly in the tails which contain rare conditions and underrepresented attributes \citep{ganapathi2022tackling}. Extending these data by generating unseen yet realistic instances can augment the downstream task to allow for novel analyses and hypothesis generation.

For downstream tasks to be representative, it is crucial that the generated samples remain realistic and reflective of the data intended for study. However, maintaining realism is a difficult task and must be finely balanced with the requirement to generate new samples instead of simply recreating those seen in the training set. In other fields where data generation is used, the same principle applies.
In state-of-the-art image generation \citep{ramesh2022hierarchical, rombach2022high} this trade-off has been finely balanced. The image quality has reached almost impeccable realism yet the models are able to create almost completely novel outputs.
In code generation and completion, the value of quality (code that compiles and suits the context) is higher than the value of novelty. This has resulted in issues with models perfectly reconstructing samples from the training set.

Text generation \citep{brown2020language}, a sequence generation task, is more similar to wearable data generation. These systems typically make use of autoregressive methods to predict the next word in the training set. The model can then be run on new input data and the next word prediction used for generation instead. Data generation for healthcare is an emerging field. Due to the potential high-risk of applications, data realism is even more of a concern than in other domains. Additionally, privacy concerns have historically limited access to large datasets to enable training of realistic generative models.

Methods for time-series generation exist in the literature. \citep{kang2020gratis} presented an approach using mixture autoregressive (MAR) models which can be configured to give the time series certain characteristics. The model was released as a shiny app where the properties can be configured. One drawback of this approach is that the specific characteristics, such as seasonal strength and stability, need to be quantified and cannot be inferred from the context, such as a medical condition. For healthcare data, \citet{norgaard2018synthetic} presented a Generative Adversarial Network (GAN) for accelerometer and exercise data. 
\citep{dash2020medical} also used GANs for generation of hospital time-series based on the MIMIC-III dataset. More recently, outside healthcare applications, \citet{srinivasan2022time} and \citet{li2022tts} have proposed a general architecture based on transformers but train it using the GAN framework. 

In this work, we focus on personal health data, specifically multi-modal resting heart rate, sleep and step data, generated by consumer wearable devices. Applications on the health domain of such data are still emerging, detection of flu and COVID-19 being one example \citep{shapiro2021characterizing, merrill2022self}. Our approach features a multi-task self-attention model for wearable activity data synthesis. 

In summary, our contributions are:
\begin{itemize}
    \item A synthetic data generator based on self-attention for wearables data
    \item Demonstration that the model can predict future activity through self-supervised learning of over 2 million activity days
    \item Evaluation of the generative model with qualitative and quantitative comparisons to genuine real-world data
\end{itemize}

\section{Data for training}
\paragraph{Dataset.} 
All models were trained and evaluated on the same set of activity data acquired using wearable FitBit trackers, collected as part of the DiSCover (Digital Signals in Chronic Pain) Project, a 1-year longitudinal study (ClinicalTrials.gov identifier: NCT03421223) \citep{lee2021discover}. The dataset contained day-level data from \(10\,000\) individuals who gave permission for use of their data for the purpose of health research. Data were collected over one year, resulting in a total of \(2\,737\,500\) person-days of activity data. The data contain three signals: resting heart rate (beats per minute), total sleep (minutes), total steps (step count). The mean age of the participants was 37.3 (SD=10.5, range: 18 to 85) with 72.15\% of participants female and primarily Non-Hispanic White (80.5\%).

\paragraph{Pre-processing.} Day level aggregates were calculated from the minute-level raw data by summing all minutes spent sleeping per day, summing all steps per day and taking the mean resting heart rate per day. Only days with \(> 80\%\) coverage were included in the analaysis. Missing data were imputed with the mean feature values per individual. Each feature was then scaled to \([0, 1]\). We then divide the year-long sequences into shorter sequences with a length of 21 days for use as inputs. Although this is much shorter than sequences used with most transformers, we keep this short for the following reason: every source sequence is of length 365, corresponding to each day in the year for an individual. If we use a larger window of, e.g., 100 we could only create three non-overlapping sequences per individual. The shorter sequence length gives us a more diverse set of samples while still capturing a representative time period on the scale of human activity (three weeks).

Although the labels are continuous values, we convert them to a one-hot encoding of \(100\) evenly-spaced bins. We do this to model the outputs as a softmax distribution. As described by \citet{van2016pixel}, this removes any assumptions about the shape of the distribution and is therefore highly compatible with neural networks and has also been used for audio-generation in Wavenet \citep{oord2016wavenet}.

\section{Model and learning}
\paragraph{Embeddings.} The three input channels (resting-heart-rate, sleep minutes and step count) are embedded in a 64 dimensional space through a learned embedding weights. As the sequences are temporally ordered, it is important to preserve their positional relationships. To do that, they are positionally encoded with learned positional weights that are added to the embedded inputs.

\paragraph{Transformer.} The embeddings are passed into a transformer \citep{vaswani2017attention} that consists only of decoder layers.
Self-attention is calculated as \( attention(Q, K, V) = softmax(QK^{T} / \sqrt{d_k}) V \). Where \(Q\), \(K\) and \(V\) are the query, key and value matrices, respectively and \(d_k\) is the dimensionality of the keys.
Decoder-only transformers have been shown to perform well in autoregressive tasks, like next-word predictions \citep{brown2020language, rae2021scaling} and joint learning of multiple tasks \citep{reed2022generalist}.
Each transformer block begins with layer-normalization to stabilize gradient updates and training. 
As this is an auto-regressive task, we ensure future information is not used by causal masking, i.e. confining each position to previous positions or the current position. This is implemented by masking the upper-right triangle of the attention weight-matrix.

Finally, each block is completed by a feed-forward network of two dense layers of dimensionality 256 with GeLU activation and dropout probability of \(0.1\) during training. We stack three of these blocks to form the core of the model, and four attention heads. It is followed with a feed-forward network to an output of three 100-unit vectors, corresponding to the three tasks and 100 bins. A softmax activation is applied to each one to obtain the logits used for loss calculation.
This results in a causally-masked multihead multi-task self-attention model that can be trained to model and forecast activity time series.

\paragraph{Loss.} As described in detail earlier, we use a softmax distribution of outputs. Then we can minimize the cross-entropy loss between the predicted and true values. We learn the three outputs (resting heart rate, daily steps and sleep minutes) jointly with separate feed-forward network heads. The individual losses are added through shake-shake regularization \citet{gastaldi2017shake}, a stochastic affine combination. The combined loss which we minimize is then defined as \[\mathcal{L}_{combined} = \sum_{i=1}^{N} \alpha_i\mathcal{L}_i\]

where \(\bm{\alpha}\) is a random vector of unit length and \(\mathcal{L}_i\) are individual losses. In our case, \(N = 3\).

\paragraph{Training.} We minimize the loss using Adam \citep{kingma2014adam} and an initial learning rate of \(10^{-3}\), reducing it by a factor of 10 every 5 epochs, with a total of 15 training epochs. The model and training were implemented in PyTorch \citep{paszke2019pytorch}, along with NumPy \citep{harris2020array} and SciPy \citep{2020SciPy-NMeth}, and visualizations in Matplotlib \citep{Hunter:2007}.

We train four different models to compare the effect of increased number of training points on the quality of generated samples. The largest model contains \(2\,029\,230\) days, which represent \(100\%\) of the available training data. We then train three smaller models with \(10\%\), \(1\%\) and \(0.5\%\) of the available training data, respectively.

\paragraph{Generating new samples}
With the autoregressive model already trained to predict next-day values, synthesizing new sequences is straightforward. We start with a prompt sequence fragment, taken from a held-out set, and input into the trained model. Then, we recursively remove the first day of the sequence and append the next-day predictions to the end. Scaling the temperature of the logits gave more consistent results for resting steps and sleep, we used temperatures of \(2\), while resting heart rate was kept with a temperature of \(1\). The three softmax distributions of the output were sampled independently to obtain the next-day value. 

\section{Results and evaluation}
We evaluate the model on four criteria. 1) The prediction accuracy of the model 2) Qualitative visualization analysis of the generated sequences 3) Quantitative evaluation of distance measures and similarity scores between real and generated sequences and 4) Comparison of real and generated sequences on a lower-dimensional manifold.

\subsection{Activity modelling}
We begin by comparing the accuracy of the next-day predictions with the ground truth real-world data. These results are highlighted in Table~\ref{tab:predictions}. Increasing the number of training samples has a strong effect, particularly on resting heart rate prediction where the mean absolute error (MAE) is reduced to \(1.21\) BPM in the case of 2 million training samples. Given only \(0.5\%\) of the data, the accuracy is far lower and increasing the number of data always results in a marked increase in accuracy. The effect of increased data has a different effect for both steps and sleep minutes. It appears that going from \(\sim20\)k days to \(\sim200\)k days has a far greater effect than the next order of magnitude, which appears to have no marked difference.

\begin{table*}[!hbtp]
\floatconts
  {tab:example-booktabs}
  {\caption{Comparison of mean absolute errors (MAE) of next-day resting heart rate (HR), sleep and steps with respect to the size of the training set. There is a marked difference in terms of accuracy as the number of training samples increases.}}
  {\begin{tabular}{rccc}
  \toprule
  \bfseries Training size &  \bfseries MAE Resting HR& \bfseries MAE Sleep & \bfseries MAE Steps \\
  \bfseries (Days) &  \bfseries (BPM) & \bfseries (Minutes) & \bfseries (Count) \\
  \midrule
  \(10\,146\) (0.5\%) & 31.9 & 135.9 & 4922\\
  \(20\,292\) (1\%) & 18.6 & 137.2 & 4444\\
  \(202\,923\) (10\%) & 3.31 & 58.6 & 2627\\
  \(2\,029\,230\) (100\%) & 1.21 & 56.2 & 2830\\
  \bottomrule
  \end{tabular}}
  \label{tab:predictions}
\end{table*}

\subsection{Visual comparisons}
\begin{figure*}[!htbp]
    \centering
        \includegraphics[width=0.80\textwidth]{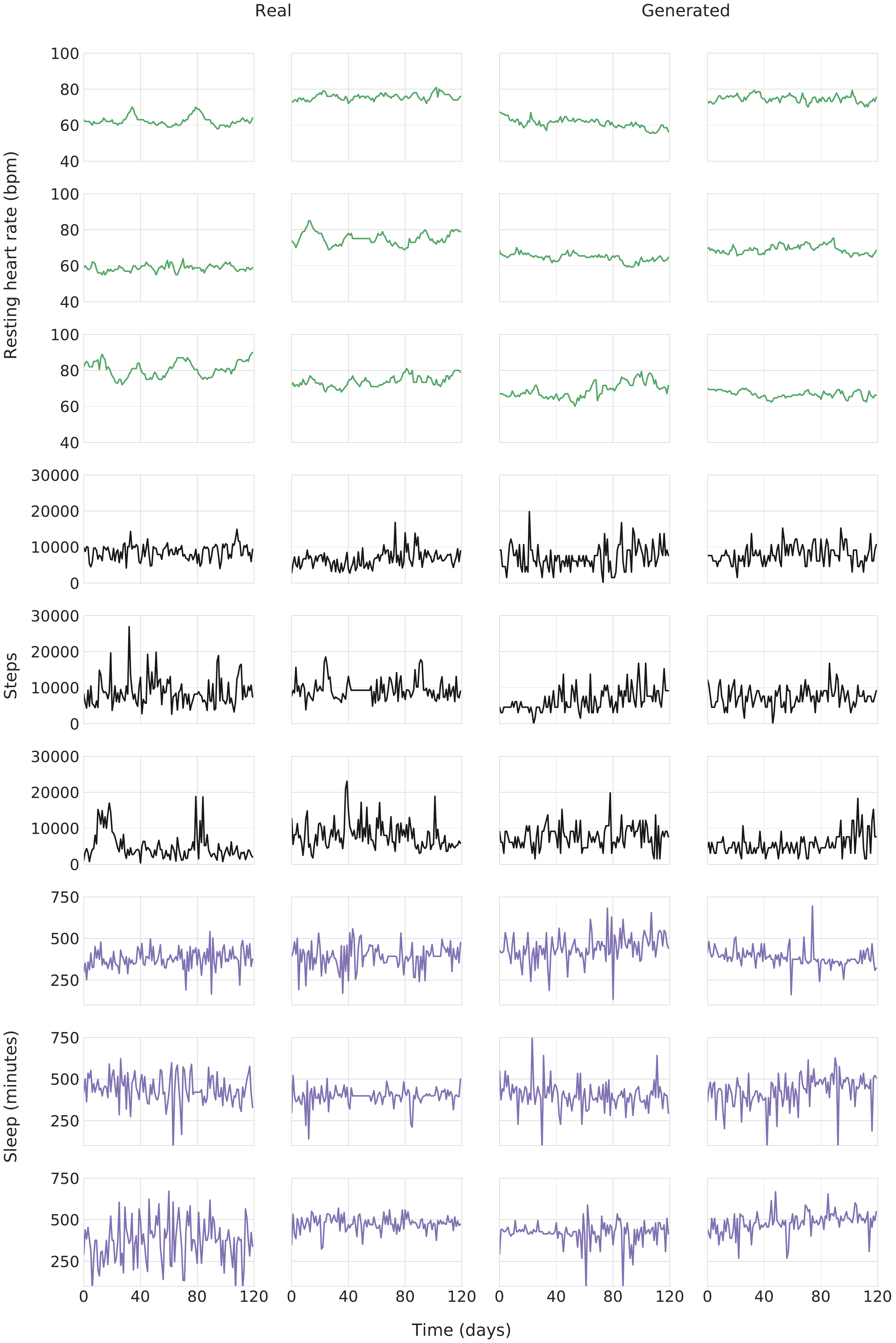}
    \caption{Comparison of real and generated wearable activity data. Each subplot represents a single individual. The two left columns show real data sequences collected from a wearable FitBit device. The two right columns show synthetic sequences generated by our model. Resting heart rate is shown in the top three rows (green), steps taken per day in the three center rows (black) and total minutes spent sleeping per day in the bottom three rows (purple).}
    \label{fig:comparison}
\end{figure*}

Next, we perform a qualitative visual comparison of the generated and real data.
In Figure \ref{fig:comparison} we highlight and compare examples of real and generated activity data across three different channels: resting heart rate, steps taken and minutes spent sleeping. We plot this over three months (120 days) to inspect both short-term and long-term trends. The generated sequences (two rightmost columns of Figure \ref{fig:comparison}) are visually similar to the real examples (two left columns). The model clearly captures the individual properties of the three different modalities. Resting heart rate remains relatively stable without spikes or clear trends. Recorded and generated steps are highly variable, with differences over orders of magnitude between consecutive days and spikes representing very-high-step days. 

\begin{figure}
  \centering
  \begin{subfigure}
    \centering
    \begin{tikzpicture}
    \begin{axis}[
        xmode=log,
        log ticks with fixed point,
        width=7cm,
        ylabel= Cosine similarity score,
        xlabel= Number of training days (millions)
    ]
    \addplot table {
    0.010146 0.6659
    0.02029230 0.7262
    0.2029230 0.7729
    2 0.8096
    };
    \end{axis}
    \end{tikzpicture}
    \caption{Mean pairwise cosine similarity measure of models trained with different training set sizes, compared with real data. Models trained with more data have more similarity with genuine data. The model trained with over 2 million days achieves a score of over \(0.810\) with the intra-similarity of real data being \(0.873\).}
    \label{fig:similarity}
  \end{subfigure}%
  \begin{subfigure}
    \centering
    \begin{tikzpicture}
    \begin{axis}[
        xmode=log,
        log ticks with fixed point,
        width=7cm,
        ylabel= Dynamic Time Warping (DTW) distance,
        xlabel= Number of training days (millions)
    ]
    \addplot[color=red,mark=o,mark options={solid,fill=red}] table {
    0.010146 248114
    0.02029230 157536
    0.2029230 51495
    2 29028
    };
    \end{axis}
    \end{tikzpicture}
    \caption{Mean pairwise dynamic time warping distance of models trained with different training set sizes, compared with real data. The mean distance from the model trained with over 2 million days to the real data is \(29\,028\) with the intra-distance of real data being \(27\,897\).}
    \label{fig:dtw}
  \end{subfigure}
\end{figure}
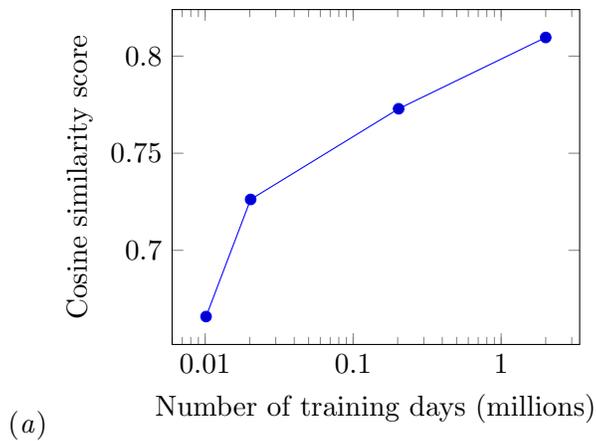

\subsection{Distance and similarity measures}

No standard collection of methods exists for scoring differences between time series. However, we make use of two common metrics: cosine similarity and dynamic time warping (DTW) distance. For cosine similarity, we follow the approach of \citet{norgaard2018synthetic} and compare the mean pairwise cosine similarity statistics between real sequences and generated ones. Where the cosine similarity statistic between two sequences \(X\) and \(Y\) is defined as their normalized dot product \( K(X,Y) = \frac{(X \cdot Y)}{ \lVert X \rVert \lVert Y \rVert}
\). The mean pairwise cosine similarity score between real sequences in the dataset is \(0.873\), providing an optimal value for this metric on the dataset. This captures the intra-dataset variation of the real data distribution.

In further analysis, we calculate the mean pairwise DTW distance \citep{bundy1984dynamic} using the DtAIdistance library \citep{wannesm2022wannesm}. The mean pairwise DTW distance in the real dataset is \(27\,897\) which provides the optimal measure for comparing the distances to the generated data.
Figure~\ref{fig:similarity} illustrates the results of this comparison. Increasing the amount of training data has a significant impact on the similarity between generated and real sequences. When only \(0.05\%\) of the total available data is used for training (\(10\,146\) days), the mean pairwise cosine similarity is \(0.666\). When \(1\%\) of the data is used, the score increases to \(0.726\), and when \(10\%\) is used, it reaches \(0.773\). The full dataset of over 2 million days yielded the best trained model with a score of \(0.810\), which is close to the intra-similarity of real data, which is \(0.873\).

In Figure~\ref{fig:dtw} we see that increasing the size of the training data results in a model that produces sequences much closer to the real data. The increase appears nearly asymptotic to the intra-distance of real data, which is \(27\,897\) compared to \(29\,028\) for data generated from the model trained on the full dataset. The agreement between the cosine similarity and the DTW distance measures provides further evidence that the model is able to capture the inherent properties of the data and generate similar sequences.

\subsection{Manifold comparisons with UMAP}
In our final set of comparisons, we compare the real and generated distributions as transformed onto a learned low-dimensional manifold using UMAP \citep{mcinnes2018umap, becht2019dimensionality}. The UMAP manifold is trained on a set of real sequences from the test set using a minimum distance of 0.1 and the cosine distance measure. A set of generated sequences from the model trained on the full dataset was then transformed onto the learned manifold.

\begin{figure*}[!htbp]
    \centering
        \includegraphics[width=0.95\textwidth]{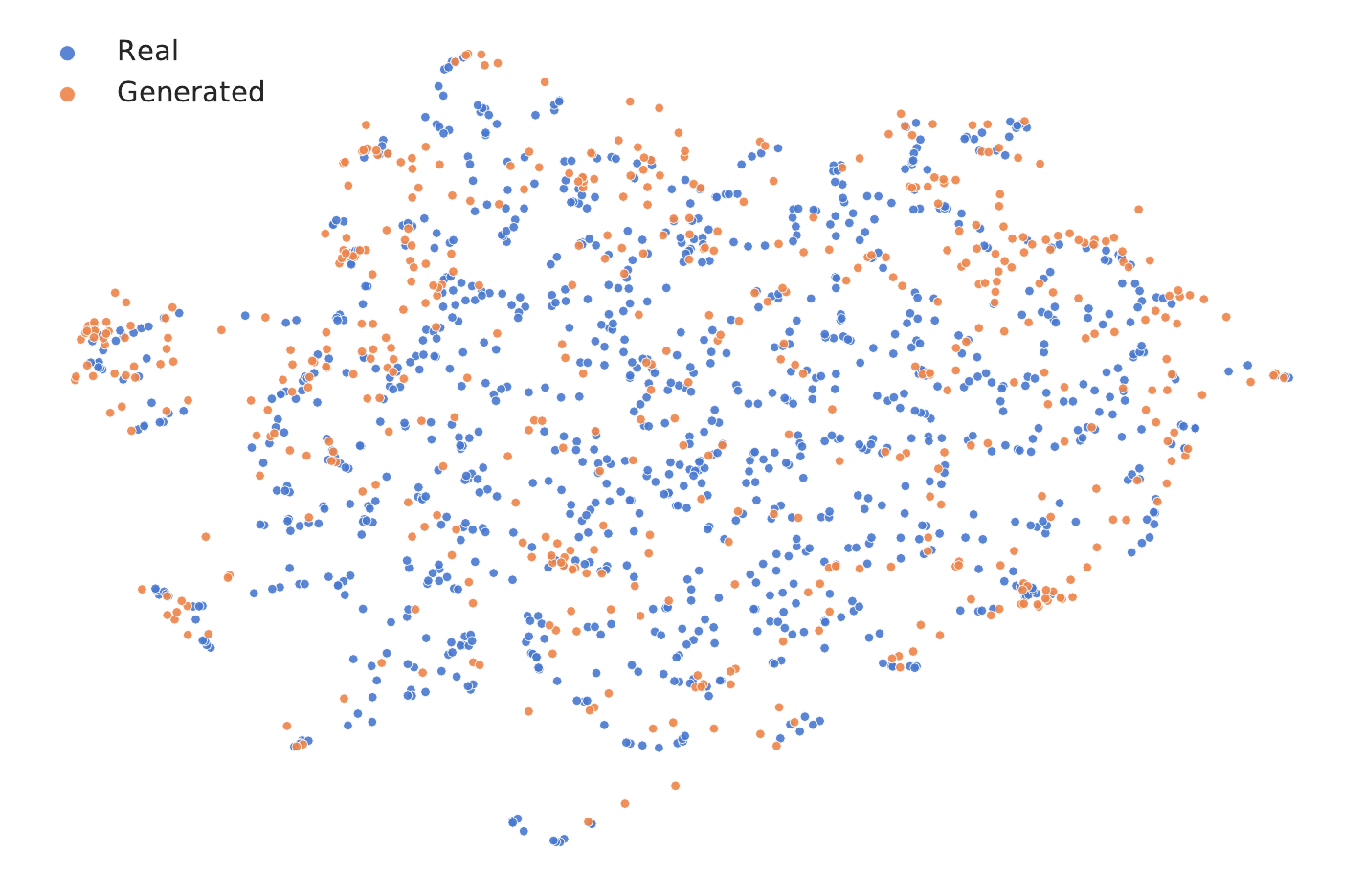}
    \caption{A UMAP manifold with real data (blue) and generated data (orange). The representation is learned with real data from the test set not seen during training. Then, generated samples are transformed and plotted simultaneously. This highlights the general landscape of the two distributions and demonstrates visually that the generated data overlaps considerably with the real data distribution.}
    \label{fig:umap}
\end{figure*}

Figure~\ref{fig:umap} visualizes this comparison. The generated samples, represented in orange, overlap very well with the real samples, represented in blue. Not only does the distribution of generated data fall within the distribution of real data, the generated data covers almost the entire surface which the real data spans. However, the densities of the two distributions appear different. One reason for this is that the generator is sampling from the correct distribution but with a biased sampling regime. Further experiments which investigate the relationship between accuracy and concentration in the distributions could help illuminate this artifact.

\section{Discussion}
We have presented a new class of activity time-series generators capable of synthesizing realistic resting heart rate, step and sleep records at the population level. It sets out necessary groundwork for conditional generators that can be controlled to output sequences with highly specific activity data properties. While synthetic activity data is an emerging field with few existing work to compare to, we note transformers have previously been used for learning from wearables data, including \citet{merrill2022self} who use minute-level data to perform influenza and COVID-19 prediction while \citet{kolbeinsson2021self} compare the performance of transformer models using different pre-training tasks.

Through our experiments we have shown that the generated data is highly similar to genuine data. The model trained on the complete set of available data (2 million days) was able to predict next-day resting heart rate with a MAE of less than 2 BPM, which is impressive. Next-day sleep was predicted to within one hour of actual sleep time and steps to within \(3\,000\). Futhermore, the DWT distance measures in addition to the mean pairwise cosine similarity demonstrated quantitatively that the generated sequences were similar to that of real data.

Synthetic wearable data has a number of applications ranging from study simulations to data visualization and quality control. Personal health research requires significant amounts of data and careful study design \citep{huang2007drug, orloff2009future}. Testing different studies and possible data collection outcomes in a simulated environment can guide study designers to set up experiments with a higher chance of success.
Similarly, synthesized data can aid in the development and testing of new analysis tools. Generated data can be modulated to allow testing of edge case and rare conditions not observed in the original real-world cohorts, without generating any privacy concerns.

Generated data could be used in privacy-sensitive research. In many environments the risk of data incidents, such as leaks or hacks during collaborations with a large number of researchers across institutions, is too great for real data testing to be viable. In such cases, data like the one presented here can be generated on-the-fly.  However, recent reports have highlighted the risks involving authorship \citep{mccormack2019autonomy, dehouche2021plagiarism} and further research into these matters is required before systems are deployed in practice.  %

One limitation of the presented approach is that generated sequences depend only on the previous 21 days. Therefore there is no direct method of interacting with the generator to request specific properties of the generated sequence, such as that of better representing a certain fitness level.
As a future research direction, we note that a slight modification to the architecture and learning process can make the model conditional, in a process similar to text-conditional image generation \citep{ramesh2022hierarchical}. It would then be easy to request a sequence with properties that the model has learned during the training process, such as age, physical fitness, and any relevant conditions such as sleep irregularities or arrhythmia. Learning these requires them to be present in the training set. 
While simpler generation approaches (e.g., sample from a statistically matched distribution) would likely give similar {\em unconditional} results to those presented here, we see the proposed architecture as groundwork for interactive generators made conditional on specific characteristics of interest. A researcher designing a study on insomnia should be able to query that ideal interactive generator for \(1\,000\) participants aged \(20\text{-}66\), have BMI \(22\text{-}30\) and half of whom sleep less than \(5\) hours per night. 

Another limitation of our work is the relatively small training dataset with respect to the general model class, transformers, which typically excel with with enormous amounts of data, and more parameters. More training data will allow us to scale up the model size even further with evidence from other domains suggesting that scale and parameter count is a powerful tool for learning richer representations \citep{brown2020language}.

Future work should focus on giving provable privacy guarantees on the generated sequences, preventing individual information from the training data to be leaked in the generated sequences \citep{mccormack2019autonomy, dehouche2021plagiarism}. Additionally, biases from the training data and other sources \citep{bender2021dangers} highlight the need for standard reporting, like model cards \citep{mitchell2019model}, for investigating and preventing risks of applied systems. 
Although the data generation is sufficiently fast for offline experiments with thousands of samples, many applications would benefit from increased efficiency and parallelisation of the generation function.

Finally, we believe that further research should go towards devising accepted benchmarks for generators such as the one presented. Unlike for images, text and code, the quality of the output cannot be easily evaluated. In addition to averages and standard deviations as proposed, a more complete suite of statistical tests should be developed to evaluate good matching, e.g., including higher order moments, tail tests, and matching in transformed spaces, such as Fourier's or Haar's.

\newpage
\section{Conclusion}
This work furthers the exploration of methods for generating synthetic personal health data. It provides researchers with the ability to craft datasets according to their needs while reducing privacy concerns, making study design more efficient and enabling the development of analysis at a faster rate. Moreover, it helps to identify issues before they affect real-world deployment. Our work adds to the existing literature on synthetic data across multiple fields and underscores the potential of generating realistic person-generated health data to enhance and improve health research.

\bibliography{refs}

\appendix

\end{document}